\documentclass[pdflatex,sn-nature]{sn-jnl}

\usepackage{graphicx}
\usepackage{amsmath,amssymb,mathtools}
\usepackage{booktabs,tabularx,array}
\usepackage{xcolor}
\usepackage{url}
\hypersetup{hidelinks,colorlinks=false,urlcolor=black,linkcolor=black,citecolor=black}
\makeatletter
\def\email#1{\global\advance\emailcnt by 1\relax
\if@corauemail
   \ifx\corrauthemail\@empty\else\g@addto@macro\corrauthemail{;\ }\fi
   \g@addto@macro\corrauthemail{\setcounter{footnote}{0}#1}
\else
   \ifx\authemail\@empty\else\g@addto@macro\authemail{;\ }\fi
   \g@addto@macro\authemail{\setcounter{footnote}{0}#1}
\fi}
\makeatother

\usepackage[mathlines]{lineno}
\makeatletter
\renewcommand\normalsize{\@setfontsize\normalsize{12bp}{15bp}}
\renewcommand\small{\@setfontsize\small{12bp}{15bp}}
\renewcommand\footnotesize{\@setfontsize\footnotesize{12bp}{15bp}}
\renewcommand\abstractfont{\reset@font\normalsize\leftskip=0pt\rightskip=0pt\parfillskip=0pt plus 1fil}
\renewcommand\abstractheadfont{\reset@font\normalsize\bfseries\titraggedcenter}
\renewcommand\keywordfont{\reset@font\normalsize\leftskip=0pt\rightskip=0pt plus0.5fill}
\renewcommand\addressfont{\reset@font\normalsize\titraggedcenter}
\renewcommand\bmheadfont{\reset@font\normalsize\bfseries\raggedright\boldmath}
\renewcommand\subsubsectionfont{\reset@font\normalsize\bfseries\raggedright\boldmath}
\renewcommand\figurecaptionfont{\reset@font\fontfamily{\rmdefault}\fontsize{12bp}{15bp}\selectfont}
\renewcommand\tablecaptionfont{\reset@font\fontsize{12bp}{15bp}\selectfont}
\makeatother
\normalsize
\newcolumntype{Y}{>{\raggedright\arraybackslash}X}
\begin{document}

\title[From internal representations to model improvement through prediction errors]{From internal representations to model improvement through prediction errors}

\author*[1,2]{\fnm{Yushi} \sur{Nakaya}}\email{nakaya@adansons.ai}
\author[1]{\fnm{Kenichi} \sur{Higuchi}}
\author[2]{\fnm{Shuichi} \sur{Ishida}}

\affil*[1]{\orgname{Adansons Corp.}, \orgaddress{\street{214 Homer Ave}, \city{Palo Alto}, \postcode{94301}, \state{CA}, \country{USA}}}
\affil[2]{\orgname{Tohoku University}, \orgaddress{\street{6-6-11-803 Aza-Aoba Aramaki, Aoba-ku}, \city{Sendai}, \postcode{980-8579}, \state{Miyagi}, \country{Japan}}}

\abstract{With limited annotation budgets, choosing which images to label determines how much a model improves. Data-selection methods that use features from a separately trained model, or scene descriptions written by vision-language models, have been successful, but those signals do not directly capture changes in the model being improved. The target model's own internal features reflect what it has learned so far and change with retraining, making them a natural cue for choosing the next training data. However, feature rarity alone does not reveal the errors that matter for performance. Here we link internal features to prediction errors and their expected impact on performance and select images for labeling and retraining without using labels for candidate images. We evaluated the method with an object detector on two datasets and two pairs of random seeds. Adding internal features improved the identification of prediction errors in 15 of 16 conditions. When performance was averaged over successive labeling rounds, the method outperformed selection based only on feature rarity in all four evaluation settings and ranked among the top two of six methods. With other conditions held fixed, performance after retraining was again higher than with rarity-based selection, even though the latter collected more errors. With longer retraining, the proposed method ranked first among six methods. These results suggest that linking a model's internal features to its errors and their effects on performance may help select training images that improve performance, thereby allowing the model's current state to guide which images are labeled next.}


\maketitle

\section*{Introduction}

The performance of supervised machine learning depends on the model size, the data selected for labeling, label accuracy, and the stage at which the data are introduced during training. In practice, experts have limited time and resources for labeling and verification, and data quality problems can persist from training to deployment \cite{sambasivan2021}. Active label cleaning selects records for review according to their likelihood of containing errors and the effort required to verify them, rather than reviewing every label, and can improve the data quality under a fixed budget \cite{bernhardt2022}. The combination of active and self-supervised learning can also improve performance while reducing the number of labels required, thereby increasing annotation efficiency \cite{geuenich2024}. Data-centric AI treats data collection, selection, labeling, and quality control as iterative processes guided by downstream model performance \cite{whang2023,mazumder2023}. For image-based models, this raises the question of which images should be selected for further labeling or inclusion in training, and how the current state of the model should inform that choice.

However, selecting images that a model has difficulty processing does not necessarily improve performance. High-loss images may include useful examples from which the model has not yet learned. However, they may also include duplicates, noisy images or labels, and examples from which the current model cannot benefit under the existing training procedure. Studies that distinguish learnability from the ability to generalize have found that substantial training effort can be spent on unnecessary or unsuitable data \cite{mindermann2022}. The value of an image also depends on model maturity: typical examples may benefit an immature model, whereas atypical examples may become more useful later in training \cite{hacohen2022}. Therefore, the value of additional images should be assessed over a learning curve rather than at a single label budget. This distinction suggests two separate tasks: identifying images on which the current model is likely to err and determining whether selected images improve the model after retraining.

Active learning identifies unlabeled data that are expected to benefit a model and selects them for labeling \cite{settles2009,ren2021survey}. We use the term \textit{selection method} to refer to the rule that determines which images should be labeled next. In deep learning, selection methods include predictive uncertainty \cite{gal2017}, core-set selection, which aims to cover the image feature space \cite{sener2018}, and BADGE, which combines output uncertainty with diversity in gradient representations \cite{ash2020}. For object detection, selection methods have used classification and localization distributions \cite{choi2021}, entropy and diversity \cite{wu2022}, consistency under image transformations \cite{yu2022cald}, and internal features combined with uncertainty \cite{yang2024}. The performance of deep active learning can depend on random seeds for image selection and model training, the initial labeled sets, annotation budgets, retraining procedures, and evaluation protocols. One method may outperform random selection under one set of conditions but not another \cite{munjal2022}. Therefore, evidence from a single pair of random seeds or a single label budget may be insufficient to establish an advantage over random sampling. We used two pairs of image-selection and model-training seeds and compared the methods over complete learning curves.

Selection methods differ in both the features they use and their relationship to the model being trained. One successful line of work selects data using representations learned separately from the target model. In the fully supervised setting of those studies, TypiClust selects typical examples from diverse clusters in a self-supervised feature space, and ProbCover selects examples to cover the distribution in such a space \cite{hacohen2022,yehuda2022}. More recently, vision-language models have been used to describe relevant detection scenarios in language and retrieve matching images for labeling, as in the AIDE data engine for autonomous driving \cite{liang2024aide}. These methods can characterize the data without relying on a target model trained on very few labels, but they rely on a separate model rather than on the behavior of the detector being trained. Neither a separately learned feature space nor a language description directly tracks changes in the target model's parameters or its remaining errors as training proceeds. An image that is distinctive in that space need not provide information that would benefit the target model at its current stage. Annotation effort may therefore be directed toward images that are unusual in the external feature space but that the target detector already handles correctly. Conversely, errors in images that appear ordinary in that space may be overlooked. Diversity-based selection can thus choose images that provide the model with little new information \cite{ash2020,sener2018}. In our comparison, this line of work is represented by $k$-center selection on features from a separately pretrained ResNet-50. We chose it because it provides a simple implementation of the shared principle of covering an external feature space, has no component tied to the target detector, and can be applied identically at every stage and to every detector.

The target model's own internal features offer a different source of information: they are computed using its current parameters and change as the model is retrained. Existing approaches already use this connection. Core-set selection measures distances between representations produced by the target network, and BADGE combines the network's internal representations with its current predictions to construct gradient embeddings \cite{sener2018,ash2020}. This dependence on the target model motivates our use of internal features: they reflect how the model currently processes each image and may therefore help reveal its specific weaknesses. For an object detector, the features can be obtained during the same inference pass that produces the predicted boxes, without a separate feature extractor. The question is how to turn this information about the current model into a useful criterion for selecting its next training images.

Using internal representations alone does not guarantee better performance. Selecting for feature-space coverage or gradient diversity, as in core-set selection and BADGE \cite{sener2018,ash2020}, does not explicitly distinguish error types and weight them by their effects on the evaluation metric. In particular, feature rarity does not directly indicate the probability of a classification or localization error, its effect on performance, or whether retraining can correct it. We hypothesize that linking internal features to the target model's prediction errors and their expected impact on performance can make those features more useful for image selection. This principle may also be applicable to other learning tasks in which error types and their effects on an evaluation metric can be defined. We tested this hypothesis for object detection.

For object detection, the target detector predicts bounding boxes and class labels for objects in each image. Errors can arise from incorrect class labels, inaccurate box locations, duplicate detections, predictions on background regions, or missed objects. These error types can affect evaluation metrics differently. The Toolbox for Identifying Detection Errors (TIDE) attributes the losses in detection performance to these error types \cite{bolya2020}. However, classifying errors based on ground truth differs from predicting errors in unlabeled images before selection. Detector confidence or class probabilities are commonly used as indicators of prediction uncertainty, but neural network probabilities are not necessarily calibrated \cite{guo2017}. In object detection, calibration error can also vary with the position and size of the predicted bounding box \cite{kuppers2020}. Thus, low confidence alone does not indicate that a prediction is erroneous, that the error is worth correcting, or that the corresponding image would be useful for further training. Certain post-processing methods use detector outputs alone to estimate whether a predicted box is correct \cite{schubert2021}. However, they do not use internal representations and reduce errors to a distinction between true and false positives. When used independently, internal features and output uncertainty do not identify errors in an image or indicate how those errors relate to performance after retraining.

In this study, we therefore use the errors made by the target model and their expected effects on performance to guide image selection. We estimated the type and probability of each prediction error from the target model's internal representations and outputs without using ground-truth annotations for the candidate images and combined these estimates with the expected effect of each error on the current evaluation metric. We did not attempt to predict whether training with an image would correct the errors associated with it. Instead, we changed only the selection method, retrained the same model using the same procedure, and evaluated the selection cue using performance over the resulting learning curve.

\begin{figure}[htbp]
\centering
\includegraphics[width=\textwidth]{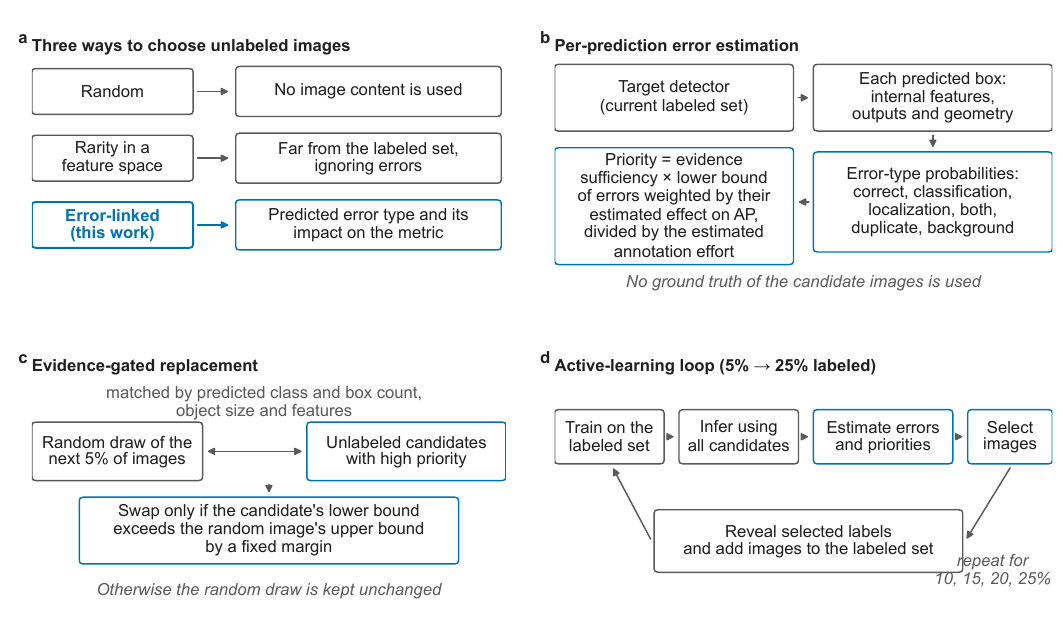}
\refstepcounter{figure}\label{fig:conceptual-overview}\par\smallskip
\begin{minipage}{\textwidth}\setlength{\parindent}{0pt}\textbf{Fig. 1 $|$ Selecting data by linking a model's internal representation to its prediction errors.} \textbf{a}, Three ways to choose unlabeled images. \textbf{b}, Six outcome probabilities per box are estimated from internal features, outputs, and geometry. The image priority is the evidence sufficiency times the lower stability bound of the error-weighted sum, obtained by resampling the predicted boxes, divided by the estimated annotation effort. Ground-truth annotations for the candidate images are withheld. \textbf{c}, A randomly drawn image is replaced by a matched candidate only when the candidate's lower bound exceeds the random image's upper bound by a fixed margin. \textbf{d}, Active-learning loop from 5\% to 25\% labeled images.\end{minipage}
\end{figure}

In the proposed method, the target detector trained on the current labeled set is fixed and runs on all unlabeled candidates. The error type and error probability of each predicted box are estimated from the internal features, final outputs, and position and size of the box. Each error-type probability is weighted by the change in average precision (AP) that would follow if the errors of that type were replaced by correct predictions, and the weighted values are then summed within the image. The sufficiency of the evidence, the stability of the estimate under resampling of the predicted boxes, and an estimate of relative annotation effort computed from the number and crowding of predicted boxes are then considered together to determine the image priority (Fig.~\ref{fig:conceptual-overview}). The selection starts from a random draw obtained using the same image-selection seed. Candidates are matched to random images with similar predicted classes, object counts, object sizes, and internal features, and a random image is replaced only when the candidate's priority is clearly judged to be higher. When the evidence is insufficient for determining priority, the random draw is retained. Annotations are revealed only after selection, and selection and retraining are repeated as the labeled fraction of candidate images increases from 5\% to 25\%.

The primary analysis used YOLOv8n with COCO 2017 and BDD100K \cite{jocher2023,lin2014,yu2020bdd}. COCO 2017 contains everyday objects, whereas BDD100K contains driving scenes. Together, they allowed us to test the selection principle across two distinct image domains. For each dataset, we used image-selection seed 11 with model-training seed 1, and image-selection seed 18 with model-training seed 0. The second pair was used to test whether the direction of the results is reproduced when both seeds change. We compared six methods: the proposed method, random sampling, $k$-center selection using features from a separate ResNet-50, BADGE adapted to object detection (hereafter BADGE), which uses internal representations and output uncertainty, output-entropy selection, and rarity-only selection based on internal representations. Within each dataset and seed condition, all methods started with the same labeled images and the same model trained on the initial 5\% of labeled images (hereafter, the 5\% model) and then increased the labeled fraction to 10, 15, 20, and 25\%. These fractions refer to detector-training images. They exclude a separate labeled set of 1,500 images that the proposed method uses to fit, tune, and evaluate its error estimator (Methods). Our primary metric was the normalized area under the five-point learning curve. We connected the mAP@[0.50:0.95] (mean average precision over box-overlap thresholds from 0.50 to 0.95, hereafter mAP) values at each point with straight line segments, integrated the resulting curve using the trapezoidal rule, and divided the area by the 0.20 width of the label range. A matched comparison from the same 10\% model to the 15\% model isolated the difference between linking internal representations to prediction errors and selecting images based on representation rarity. We report both the number of true errors in the selected images and the mAP after retraining. Thus, the comparison tests whether collecting more errors coincides with improved performance. A supplementary evaluation used Faster R-CNN with image-selection seed 11 and model-training seed 1 to compare the proposed method, BADGE, and rarity-only selection using the corresponding 5\% model for each dataset. We also tested the sensitivity to the training time in a BDD100K experiment that trained every stage for 40 epochs from the same 5\% model.

We show that internal features identified the detector's erroneous predictions more accurately than an estimator without them in 15 of 16 conditions. We also demonstrate that error-linked selection outperforms rarity-only selection in a matched comparison, although the latter collected more true errors. The proposed method ranked among the top two of the six methods in all four main conditions and first of six when every stage was retrained for 40 epochs. Under the conditions examined, the errors of the current model and their effects on the evaluation metric were more useful cues for choosing training images than the rarity of the internal representation.

\section*{Results}

\subsection*{Internal features improved identification of the detector's erroneous predictions}

\begin{figure}[htbp]
\centering
\includegraphics[width=\textwidth]{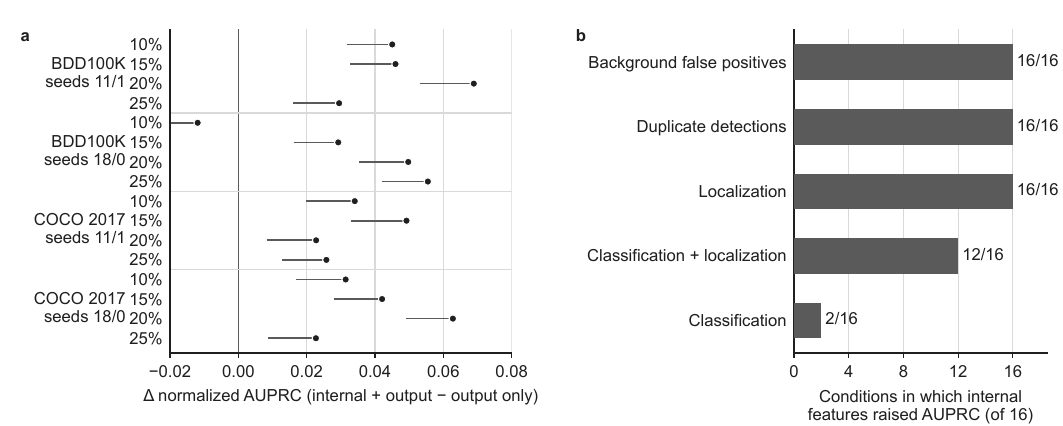}
\refstepcounter{figure}\label{fig:error-ranking}\par\smallskip
\begin{minipage}{\textwidth}\setlength{\parindent}{0pt}\textbf{Fig. 2 $|$ Error identification with and without internal features.} \textbf{a}, Difference in the normalized area under the precision--recall curve (AUPRC), averaged over error types, between the full and output-only estimators in 16 conditions: two datasets, two pairs of image-selection and model-training seeds, and four selection stages at 10 to 25\% labeled. The horizontal lines show the one-sided 90\% lower bound from the image-level resampling. The 375 evaluation images were not used for training or tuning. \textbf{b}, Number of conditions, out of 16, in which internal features increased AUPRC, by error type.\end{minipage}
\end{figure}

Adding internal features improved the identification of erroneous detector predictions for candidate images, without using ground-truth annotations. We evaluated this ability using a reserved set of 375 images with ground truth annotations. These images were not used for training or tuning. Therefore, the actual outcome of each predicted box served as a reference. An estimator that used internal features, outputs, and box information was compared with an estimator that only used outputs and box information. The ability of each estimator to identify erroneous predictions was measured using the area under the precision--recall curve (AUPRC), corrected for the prevalence of each error type, and averaged over the types. Across the 16 conditions (two datasets, two pairs of image-selection and model-training seeds, and four selection stages), the estimator with internal features scored higher in 15 of the 16 conditions (Fig.~\ref{fig:error-ranking}). In those same 15 conditions, the lower bound of the one-sided 90\% interval obtained by resampling the images was above zero. In terms of error type, internal features increased AUPRC under all 16 conditions for background false positives, duplicate detections, and localization errors. The addition of internal features increased AUPRC under 12 conditions for combined classification and localization errors, and under only two conditions for classification errors.

Our main aim was to identify likely errors for image selection rather than to improve probability calibration (Fig.~\ref{fig:error-ranking}). Adding internal features reduced the Brier score in 4 of the 16 conditions, the negative log-likelihood in 5, and the adaptive expected calibration error in 3 (Supplementary Fig.~1). These measures assess the error probability estimates themselves. Subsequent analyses assess their use in image selection.

\subsection*{Linking internal representations to prediction errors improved whole-curve performance}

The normalized area under the learning curve (AULC) represents the average performance as the labeled fraction increases from 5\% to 25\%. Within each dataset and seed condition, the six methods started from a shared 5\% model whose weights and model state were verified as identical using SHA-256 hashes. The mAP values of the shared 5\% models were 11.290 and 10.732 on BDD100K for image-selection seed 11 with model-training seed 1 and for image-selection seed 18 with model-training seed 0, respectively, and 13.528 and 13.268 on COCO 2017 for the same two seed pairs.

\begin{table}[htbp]
\refstepcounter{table}\label{tab:main-yolo-six-method}\noindent\begin{minipage}{\textwidth}\setlength{\parindent}{0pt}\textbf{Table 1 $|$ Whole-curve performance of six selection methods in the YOLOv8n main analysis.} Normalized area under the learning curve, starting from one shared 5\% model. Units are percentage points of mAP@[0.50:0.95]. Seeds are given as image-selection seed/model-training seed. Rank is the rank of the proposed method among the six methods. Differences reported in the text were computed from unrounded values.\end{minipage}\par\medskip
\footnotesize
\setlength{\tabcolsep}{2pt}
\begin{tabular*}{\textwidth}{@{\extracolsep\fill}llrrrrrrc@{}}
\toprule
Dataset & Seeds & \shortstack{Pro-\\posed} & \shortstack{Ran-\\dom} & \shortstack{External\\$k$-center} & BADGE & \shortstack{Output\\entropy} & \shortstack{Rarity\\only} & Rank \\
\midrule
BDD100K & 11/1 & 13.929 & 13.308 & 13.829 & 12.803 & 12.809 & 12.881 & 1st \\
BDD100K & 18/0 & 13.248 & 13.182 & 13.707 & 12.736 & 12.298 & 12.134 & 2nd \\
\shortstack[l]{COCO\\2017} & 11/1 & 15.839 & 15.959 & 14.707 & 15.664 & 15.484 & 15.228 & 2nd \\
\shortstack[l]{COCO\\2017} & 18/0 & 15.735 & 15.541 & 14.346 & 15.336 & 14.887 & 14.850 & 1st \\
\botrule
\end{tabular*}
\end{table}

\begin{figure}[htbp]
\centering
\includegraphics[width=\textwidth]{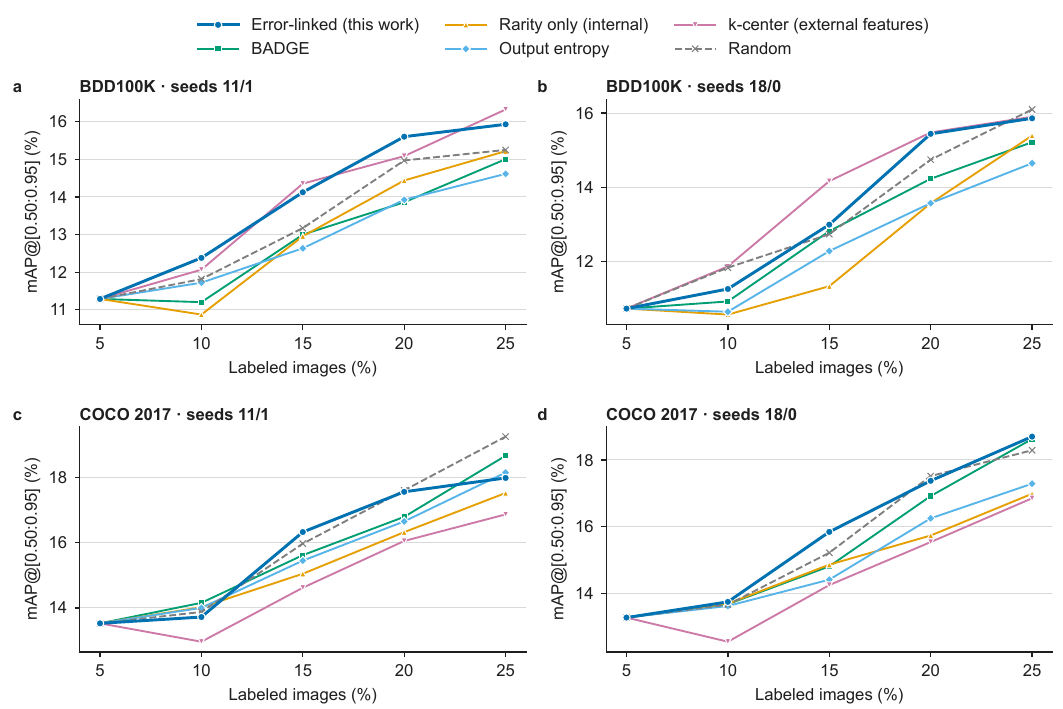}
\refstepcounter{figure}\label{fig:harmonized-active-learning-curves}\par\smallskip
\begin{minipage}{\textwidth}\setlength{\parindent}{0pt}\textbf{Fig. 3 $|$ Learning curves of six selection methods starting from a shared 5\% model (YOLOv8n).} \textbf{a, b}, BDD100K and \textbf{c, d}, COCO 2017. Left column: image-selection seed 11 and model-training seed 1. Right column: image-selection seed 18 and model-training seed 0. Vertical axis: mAP@[0.50:0.95] (mean average precision over box-overlap thresholds from 0.50 to 0.95) on the official validation images. Horizontal axis: fraction of labeled images. The six methods in each condition share a 5\% model whose weights and model state match down to the SHA-256 hashes, so differences from 10\% onward arise from image selection and subsequent retraining. The labeled fractions count detector-training images and exclude the separate 1,500-image labeled set used by the proposed method for error estimation.\end{minipage}
\end{figure}

The proposed method ranked first, second, second, and first under the four conditions (Table~\ref{tab:main-yolo-six-method}, Fig.~\ref{fig:harmonized-active-learning-curves}). Its margin over rarity-only selection ranged from +0.611 to +1.113 points, and the conditional 95\% intervals from 10,000 matched resamples of the evaluation images were above zero for all four conditions. The margin over output-entropy selection ranged from +0.355 to +1.120 points, with all four intervals above zero. The margin over BADGE ranged from +0.175 to +1.126 points. The point estimates were positive in all four conditions. However, the conditional 95\% interval was above zero in three conditions and included zero for COCO 2017 with image-selection seed 11 and model-training seed 1.

Compared with external-feature $k$-center selection, the point estimate of the proposed method was higher in three of the four conditions and 0.459 points lower on BDD100K with image-selection seed 18 and model-training seed 0. Compared with random sampling, the normalized area under the learning curve of the proposed method was higher in three of the four conditions and 0.120 points lower on COCO 2017 with image-selection seed 11 and model-training seed 1. The interval for the difference from random sampling was above zero on BDD100K with image-selection seed 11 and model-training seed 1, and on COCO 2017 with image-selection seed 18 and model-training seed 0, whereas it included zero in the other two conditions. A sensitivity analysis recalculated the 20 pairwise differences between the proposed method and the five comparators across the four experimental conditions using only the 10\%, 15\%, 20\%, and 25\% labeled-data points and excluding the shared 5\% starting point. The direction of all 20 differences remained unchanged.

\subsection*{Selection linking internal representations to prediction errors outperformed rarity-only selection}

\begin{figure}[htbp]
\centering
\includegraphics[width=\textwidth]{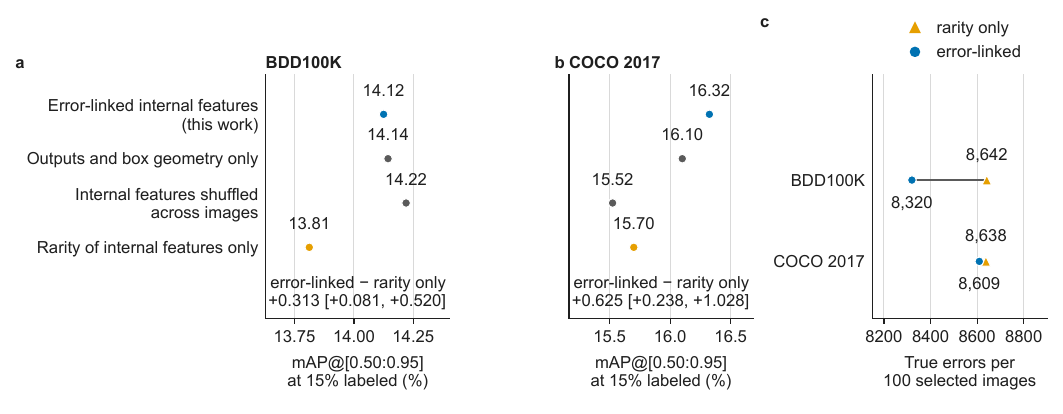}
\refstepcounter{figure}\label{fig:mechanistic-intervention}\par\smallskip
\begin{minipage}{\textwidth}\setlength{\parindent}{0pt}\textbf{Fig. 4 $|$ Matched comparison of four image-selection criteria.} One matched comparison from the common 10\% model to 15\% labeled images with image-selection seed 11 and model-training seed 1. The starting model, candidate images, replacement cap, training, and evaluation were identical. Only the pre-specified priority criterion differed (four variants). \textbf{a}, BDD100K and \textbf{b}, COCO 2017: mAP@[0.50:0.95] after retraining. Brackets provide a 95\% interval of error-linked minus rarity-only from 10,000 resamples of the evaluation images, with the trained models fixed. \textbf{c}, True errors contained in the selected images per 100 images.\end{minipage}
\end{figure}

At the 15\% labeled-data level with image-selection seed 11 and model-training seed 1, we held the starting model, candidate images, maximum number of replaced images, training conditions, and evaluation method fixed and changed only the criterion used to prioritize images. The replacement cap was 1,684 images on BDD100K and 2,894 on COCO 2017, half of the images added at that stage. The proposed method replaced 1,684 and 264 images, respectively. The numbers of images replaced under the other three criteria were not retrieved from the run records. The proposed method exceeded rarity-only selection by 0.313 points on BDD100K, with a conditional 95\% interval of +0.081 to +0.520 from resampling the evaluation images with the trained models fixed, and by 0.625 points on COCO 2017 with an interval of +0.238 to +1.028 (Fig.~\ref{fig:mechanistic-intervention}).

The rarity-only selection method selected images containing more true errors: 8,641.627 per 100 images on BDD100K and 8,637.640 on COCO 2017, compared with 8,319.952 and 8,609.190, respectively, for the proposed method. However, the proposed method achieved a higher mAP after retraining. Among all four criteria at the 15\% labeled-data level, the proposed method ranked highest on COCO 2017. For BDD100K, it was 0.019 points below the output-based criterion and 0.095 points below the criterion with internal features shuffled across images. These results support two limited conclusions: (a) solely collecting more true errors does not explain the performance after retraining, and (b) the proposed method differs from the rarity-only selection method by using predicted error types and their expected effects on the evaluation metric rather than feature rarity alone.

\subsection*{Replacement behavior remained consistent when the image-selection and model-training seeds changed}

For the proposed method and BADGE, we retrieved the selected image IDs and selection summaries saved at every cycle and verified against a sealed hash manifest for eight experiments comprising two datasets, two methods, and two seed conditions (image-selection seed 11 with model-training seed 1, and image-selection seed 18 with model-training seed 0).

Across all cycles, the proposed method replaced 6,703 images from the random draw on BDD100K with image-selection seed 11 and model-training seed 1 and 6,676 with image-selection seed 18 and model-training seed 0. The replacement rate was approximately 49--50\% per cycle under both conditions. On COCO 2017, it replaced 1,193 and 1,146 images, or approximately 4--6\% per cycle. Thus, under both seed conditions, the method found many candidates that met the replacement criterion for BDD100K but retained most of the random draw on COCO 2017.

The Jaccard index between the selected sets of the proposed method and BADGE ranged from approximately 0.026 to 0.038 at each stage. These methods selected largely different images, and they used different selection rules and parent models after the first retraining stage. The comparison describes the consistency of the proposed method's replacement behavior rather than the isolated effect of the error information.

\subsection*{The proposed method achieved the highest whole-curve performance among six methods under 40-epoch retraining}

\begin{figure}[htbp]
\centering
\includegraphics[width=\textwidth]{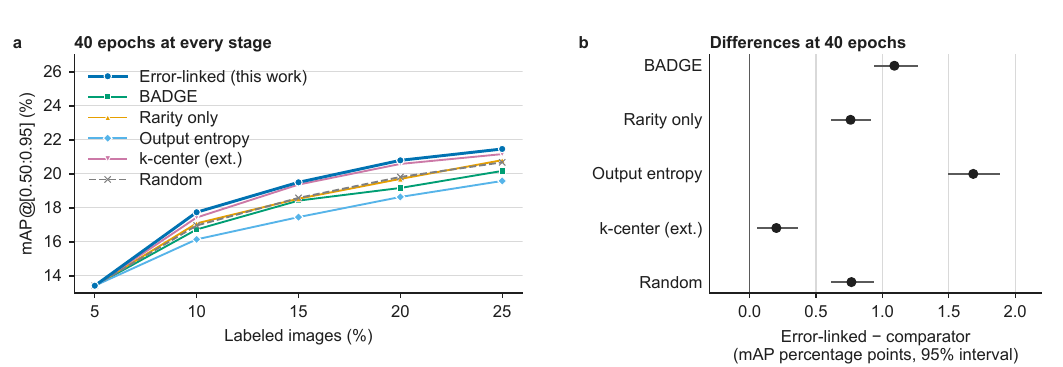}
\refstepcounter{figure}\label{fig:forty-epochs}\par\smallskip
\begin{minipage}{\textwidth}\setlength{\parindent}{0pt}\textbf{Fig. 5 $|$ Learning curves and method differences with 40 epochs of retraining at every stage.} YOLOv8n on BDD100K with image-selection seed 11 and model-training seed 1. All methods shared identical 5\% model weights. \textbf{a}, Learning curves with 40 epochs at every stage for the five prespecified methods and for rarity-only selection, which was added in a comparison planned after the five-method results had been observed. \textbf{b}, Difference in normalized area under the learning curve between the proposed method and each comparator, with conditional 95\% intervals from 10,000 resamples of the evaluation images with the trained models fixed.\end{minipage}
\end{figure}

To examine the sensitivity to training time, we restricted the experiment to YOLOv8n, BDD100K, image-selection seed 11, and model-training seed 1 and trained each of the five stages for 40 epochs. The prespecified scope of this sensitivity analysis comprised five methods: the proposed method, BADGE, $k$-center selection on external features, random sampling, and output-entropy selection. Rarity-only selection, the sixth method of the main analysis, was evaluated under the same 40-epoch schedule from the identical 5\% model in an additional comparison planned after the results of the original five methods had been observed. The normalized AULC values calculated from the official mAP values at 5\%, 10\%, 15\%, 20\%, and 25\% labeled data were 18.874 for the proposed method, 18.672 for the external-feature $k$-center, 18.113 for rarity-only selection, 18.107 for random sampling, 17.783 for BADGE, and 17.189 for output-entropy selection. The corresponding differences between the proposed method and the five comparators were +0.201, +0.760, +0.767, +1.090, and +1.684 points, respectively (Fig.~\ref{fig:forty-epochs}, Supplementary Table~10). Rarity-only selection and random sampling differed by only 0.007 points.

The conditional 95\% intervals from 10,000 resamples of the evaluation images with the trained models fixed were +0.062 to +0.357 points for the difference from the external-feature $k$-center, +0.620 to +0.905 for rarity-only selection, +0.621 to +0.930 for random sampling, +0.946 to +1.260 for BADGE, and +1.497 to +1.876 for output-entropy selection. All five were above zero (Fig.~\ref{fig:forty-epochs}). When the shared 5\% point was excluded, the differences calculated from the 10\%, 15\%, 20\%, and 25\% labeled-data points remained positive at +0.217, +0.904, +0.893, +1.285, and +1.980.

The six methods started from the same 5\% model, with identical weights, model state, and initial labeled images. This model had an official mAP of 13.424. Therefore, the differences from 10\% onward cannot be attributed to differences in training before the image-selection methods diverged.

\section*{Discussion}

We selected training images based on the target model's predicted errors and their expected effects on the evaluation metric rather than on image rarity or error counts alone. The internal features identified likely errors more accurately in 15 of the 16 conditions. We then evaluated the selected image sets using learning curves that began with a shared 5\% model, together with a matched comparison at the 15\% labeled-data level in which only the selection criterion was changed. In the whole-curve comparison, the proposed method outperformed the rarity-only selection method under all four conditions. It also ranked highest among the six methods on BDD100K when every stage was trained for 40 epochs.

In the supplementary evaluation using Faster R-CNN, the proposed method, BADGE, and rarity-only selection started from the same 5\% model within each dataset (Supplementary Table~9). The proposed method ranked third on BDD100K and first on COCO 2017. Its relative performance therefore differed between the datasets.

In the main analysis using YOLOv8n, all methods started from the same 5\% model within each dataset and seed condition. The proposed method ranked first, second, second, and first across the four conditions. It outperformed the rarity-only and output-entropy selections in all four conditions, with all conditional 95\% intervals above zero. Its margin over BADGE was positive in all four conditions, with an interval above zero in three conditions. These consistently positive point estimates relative to comparators that use internal representations support linking those representations to prediction errors and their expected effects on the evaluation metric. Compared with external-feature $k$-center selection and random sampling, the point estimate for the proposed method was higher in three of the four conditions, and lower in one. The sensitivity analysis based only on the 10\%, 15\%, 20\%, and 25\% labeled-data points left the direction of these 20 comparisons unchanged.

In the BDD100K experiment with 40 epochs at every stage, the proposed method ranked highest among the six methods that began with the same 5\% model. The differences from the five comparators remained positive after excluding the shared starting point. Under the main retraining schedule for the same dataset and seed pair, the proposed method had a higher point estimate than each comparator, and these differences remained positive with longer training. The difference from the external-feature $k$-center was +0.101 points under the main schedule, with a conditional 95\% interval that included zero ($-0.015$ to +0.205, Supplementary Table~6), and +0.201 points with 40 epochs, with an interval that excluded zero (+0.062 to +0.357). Both intervals were obtained by resampling the evaluation images with the trained models held fixed. The difference between these estimates does not establish whether longer training increases the gap between the methods. Instead, the experiment showed that the ordering was retained under a longer schedule.

The scores with internal features singled out erroneous predictions more accurately under almost all evaluation conditions, most clearly for background false positives, duplicate detections, and localization errors. However, on the Brier score, negative log-likelihood, and expected calibration error, this approach did not consistently surpass the scores that used outputs and box information alone. Therefore, internal features improved the identification of which predictions were likely to be wrong but did not demonstrably improve the accuracy of the predicted probability values. As the proposed method determines replacements by comparing these stability bounds, probabilities that are lower than the truth render replacement less likely. The method handles this mismatch using temperature scaling, the weight given to internal features, the number of similar calibration examples, and the stability bounds. Making the predicted probabilities more accurate is an improvement separate from the ranking gain and is left for future work.

Increasing diversity in the internal-representation space and improving detection performance after retraining are distinct objectives. In the matched comparison at the 15\% labeled-data level with image-selection seed 11 and model-training seed 1, the rarity-only selection method chose images containing more true errors. However, the proposed method had a higher mAP. Thus, collecting more true errors alone does not explain performance after retraining. Among the four criteria, the proposed method achieved the highest mAP on COCO 2017, whereas on BDD100K, its mAP was slightly lower than that of the output-based criterion and the criterion with internal features shuffled across images. Therefore, the independent effect of each component cannot be separated in this comparison.

The verified selection records showed that the proposed method replaced approximately half of the random draw on BDD100K and only approximately 4--6\% on COCO 2017 under both seed conditions. This dataset-specific replacement pattern was consistent across the two seed pairs, indicating that the method retained the random draw when evidence for replacement was insufficient.

The performance of the proposed method relative to the external-feature $k$-center and random sampling methods varied by dataset and seed condition. The proposed method achieved a higher normalized AULC than external-feature $k$-center selection under both COCO 2017 conditions and on BDD100K with image-selection seed 11 and model-training seed 1, but a lower AULC on BDD100K with image-selection seed 18 and model-training seed 0. It achieved a higher normalized AULC than random sampling under both BDD100K conditions and on COCO 2017 with image-selection seed 18 and model-training seed 0, but lower AULC on COCO 2017 with image-selection seed 11 and model-training seed 1. The four conditions were too few to attribute these differences to dataset or seed effects. The combination of error-based selection with diversity-based selection across the entire image collection remains to be tested.

This study had five limitations. First, only two conditions with different image-selection and model-training seeds were used to examine whether the direction of the results was reproduced. This is not sufficient to estimate the distribution over other seeds. The 95\% intervals for each setting express only the variation resulting from changing the evaluation images with the trained models fixed, and not the variation from repeating training or image selection. Second, the detector in the main analysis was YOLOv8n, and the supplementary comparison with Faster R-CNN was limited to one pair of image-selection and model-training seeds, two datasets, and three methods that use the internal representation. Random sampling, output-entropy selection, and external-feature selection were not evaluated with Faster R-CNN, and generalization to other detectors, datasets, or distribution shifts was not tested. Third, missed objects have no predicted box and hence no internal representation. Therefore, they are not included directly in the box-level error estimate. Fourth, adding internal features helped identify predictions that were likely to be errors but did not demonstrably improve the accuracy of the predicted probabilities. Fifth, savings in labeling and GPU time could not be established from the present results.

The value of the target model's internal representations lies not only in their rarity but also in their relationship to the errors made by the current model. Our results supported this design under the conditions tested in this study. The proposed method outperformed random sampling and external-feature selection under three of the four conditions. Consequently, we do not claim an advantage in every setting.

The proposed selection principle is not inherently specific to object detection. It requires an internal representation of each prediction, defined error types, and an estimate of the effect of each type on the evaluation metric. We chose object detection because its error types and their effects on the evaluation metric can be systematically characterized \cite{bolya2020}. Thus, these requirements can be specified concretely. The same procedure may be applicable to image classification, segmentation, or sequence prediction whenever the error types and their effects on the metric can be defined. The evidence for the effectiveness of the method proposed in this study is limited to the YOLOv8n main analysis on two object-detection datasets and the restricted Faster R-CNN comparison. Its effectiveness on other tasks remains to be tested.

Overall, the results indicate that information about the current prediction errors of the target model and their expected effects on the evaluation metric can provide a useful basis for selecting images for annotation and retraining. By linking the target model's internal representations to prediction error types and their expected metric effects, the proposed method yielded higher post-retraining performance than rarity-only selection under the conditions examined. The results also show that collecting more true errors does not by itself explain the performance after retraining, and that the value of a selection method should be assessed over several labeled-data levels rather than at a single annotation budget. For BDD100K, the proposed method also performed better than all five comparators when the retraining was extended to 40 epochs at every stage. Combining error information with the diversity of the full image collection remains a direction for future study. The selection step of the proposed method requires one inference pass over the candidate images and the processing of a 1,500-image calibration set without an external model or additional detector training.

\section*{Methods}

\subsection*{Overview of the study design}

The exploratory experiments used to develop the method and the comparisons reported herein were performed using different data splits, image-selection seeds, and model-training seeds. The comparators, data splits, initialization, evaluation metrics, and evaluation procedures were fixed for the reported comparisons, and the identifiers for each run's configuration and artifacts were recorded.

The main analysis employed YOLOv8n, COCO 2017, and BDD100K. The image-selection seed determines the initial labeled set and the random draw at each stage. The model-training seed determines the weight initialization, data order, and stochastic operations during training. Two conditions were used: image-selection seed 11 with model-training seed 1, and image-selection seed 18 with model-training seed 0. The seed values were fixed before the reported comparisons were run and carry no special meaning. The second pair was used to check whether the direction of the results was reproduced when both seeds changed. As both seeds changed simultaneously, their separate effects were not identified. Six methods were compared for each dataset and seed condition: the proposed method, random sampling, $k$-center selection on external features, BADGE, output-entropy selection, and rarity-only selection based on the internal representation. Within each condition, the six methods shared the same initial labeled set and a 5\% model whose identical weights were verified using SHA-256. They then proceeded to the 10\%, 15\%, 20\%, and 25\% labeled-data stages. The matched comparison, in which only the selection criterion changed in a single step from 10\% to 15\%, was treated as an additional mechanism analysis with image-selection seed 11 and model-training seed 1, separate from the whole-curve comparison.

A supplementary comparison with a different detector used Faster R-CNN with a ResNet-50 backbone and a feature pyramid. The comparison was restricted to COCO 2017 and BDD100K, image-selection seed 11 with model-training seed 1, and three methods: the proposed method, BADGE, and rarity-only selection. This restriction reflects the scope of our claim, which concerns comparisons with existing methods that use internal representation, as well as the available computing budget, which allowed only these methods to be rerun from a shared 5\% model. Random sampling, output-entropy selection, and external-feature $k$-center selection were not run from the same 5\% model for Faster R-CNN, and their comparisons were not part of the Faster R-CNN conclusions. The three methods for each dataset shared the same initial labeled set and 5\% model. Data splits, label budgets, official evaluation rules, and the definition of each selection method were identical to those for YOLOv8n, whereas the detector architecture, initial weights, layers from which internal features were taken, and training implementation were fixed for Faster R-CNN. Faster R-CNN with a ResNet-50 backbone and feature pyramid was trained with stochastic gradient descent (learning rate 0.005, momentum 0.9, weight decay 0.0005) and a learning-rate schedule that reduced the learning rate by a factor of 10 after three epochs and restarted at each stage, with an effective batch size of 16, the same epoch schedule as YOLOv8n, and no data augmentation. This supplementary comparison was used to check whether the principle of linking the internal representations to prediction errors was also observed with a different detector architecture. It was not a controlled comparison in which only the detector architecture was changed.

\subsection*{Datasets and splits}

COCO 2017 is a general object detection dataset with 80 object classes in everyday scenes \cite{lin2014}. BDD100K contains driving scenes captured at different times of day and under different weather and road conditions and uses 10 object classes \cite{yu2020bdd}. In COCO 2017, 115,787 images were selected as candidates for labeling. In BDD100K, 67,363 of the 69,863 training images with detection annotations remained after excluding the images reserved for error estimation and internal validation and were used as candidate images. Image IDs, mappings between annotations and classes, and the SHA-256 hash of each split were recorded.

From the training images of each dataset, 1,500 images were set aside for error probability estimation and 1,000 for internal validation. The former were stratified by class occurrence, object count, and object size. They were further divided into 750 fit images, 375 tuning images, and 375 evaluation images for assessing the error estimator. All boxes from a given image were assigned to the same split. The evaluation images were not used to refit the error probability model or to tune the thresholds or selection methods. The internal validation set was used only to assess the baseline performance and convergence of the target detector and to apply prespecified early stopping. The official validation set was reserved for final evaluation. It was not used for image selection, threshold tuning, early stopping, method choice, or selection of models for evaluation.

Because COCO 2017 and BDD100K provide ground-truth annotations, this study was conducted as a retrospective simulation of pool-based active learning. For each image-selection seed, 5\% of the candidate images formed the initial labeled set and the remainder formed an unlabeled pool. Four additional batches, each comprising 5\% of the candidate images, were then added to the labeled set, with evaluations at 5\%, 10\%, 15\%, 20\% and 25\%. The candidate set at each stage comprised all eligible images except those already labeled, error-estimation images, internal validation images, and images selected at earlier stages. The ground truth of a candidate image was revealed only after the image was selected.

\subsection*{Object detectors and initialization}

The main analysis used YOLOv8n, a light single-stage detector of the YOLOv8 family \cite{jocher2023} with an input resolution of 640 and an effective batch size of 16, including gradient accumulation. The implementation environment used Ultralytics 8.4.101, PyTorch 2.13.0, and torchvision 0.28.0. For initialization, only the weights of the YOLOv8n backbone pretrained on ImageNet classification whose structure, layer names, and array shapes matched the target exactly were loaded. The feature fusion and detection head parts were initialized anew. The initialization record lists the weights that were and were not loaded, any shape mismatches, and the hash of the original weight file. It also confirms that no pretrained weights were loaded into the detection head.

For each combination of dataset, image-selection seed, and model-training seed, all methods shared the initial labeled set, candidate set, optimizer, learning rate schedule, data augmentation, input resolution, effective batch size, and evaluation rules. All methods proceeded from a model trained on 5\% of the labeled data to models trained on 10\%, and the SHA-256 hashes of the shared weights and labeled set were recorded. YOLOv8n was trained with stochastic gradient descent (initial learning rate 0.01, linear decay to 0.0001 within each stage, momentum 0.937, weight decay 0.0005, three warm-up epochs) with mosaic augmentation in every epoch, HSV jitter, translation of up to 10\%, scaling of up to 50\%, and horizontal flipping with probability 0.5. The first stage was trained for 20 epochs. Stages two to five were trained for 6 epochs each, and the final model of each stage was evaluated. The learning-rate schedule restarted at each stage, whereas the weights continued from the previous stage. From stage two onward, each method resumed training from its final model of the previous stage. The official validation set was not used during training. The full-label reference runs monitored only mAP@[0.50:0.95] on the internal validation set and stopped between 20 and 200 epochs when no improvement of at least 0.001 was observed for 20 consecutive epochs. The model that first reached the best value served as the reference. The model from the final epoch was also saved.

To assess sensitivity to training time, the supplementary experiment used YOLOv8n and BDD100K with image-selection seed 11 and model-training seed 1. All five stages were trained for 40 epochs. For the five prespecified methods (proposed method, random sampling, external-feature $k$-center, BADGE, and output-entropy selection), the data splits, initially labeled images, optimizer, input resolution, effective batch size, and official evaluation rules were the same. The five methods proceeded from the same model trained on 5\% of the labeled data to models trained on 10\% of the labeled data. Matching SHA-256 hashes for the initial model weights and labeled set were confirmed. Thereafter, each resumed training using its own final model from the previous stage. Rarity-only selection was added to this experiment after the results of the five methods had been observed. It was run as one additional trajectory under the same conditions and schedule, starting from the same 5\% model (checkpoint and model-state hashes verified), with the same selection rule as in the main analysis and without the replacement rules of the proposed method. Its official evaluation used the same evaluation protocol, images, and ground truth as the five methods, and the paired bootstrap was recomputed with the five-method values unchanged.

\subsection*{Data selection methods compared}

All six methods listed in Table~\ref{tab:methods} were compared for the two seed conditions. Each method added the same number of images at each stage. The ground-truth annotations of the candidate and official validation images, as well as performance on the official validation images, were not used for image selection. The supplementary Faster R-CNN comparison used the proposed method, BADGE, and rarity-only selection. The labeled fractions included only images used to train the detector. The proposed method, and the two matched-comparison criteria derived from it (outputs and box information only, and internal features shuffled across images), additionally used the separate 1,500-image labeled set to fit, tune, and evaluate the error estimator. Random sampling, external-feature $k$-center selection, BADGE, output-entropy selection, and rarity-only selection did not use that set.

\begin{table}[htbp]
\refstepcounter{table}\label{tab:methods}\noindent\begin{minipage}{\textwidth}\setlength{\parindent}{0pt}\textbf{Table 2 $|$ Data selection methods compared in the main analysis.} For each method, the information it receives and its selection rule are listed.\end{minipage}\par\medskip
\footnotesize
\begin{tabularx}{\textwidth}{@{}p{0.19\textwidth}p{0.20\textwidth}Y@{}}
\toprule
Method & Information used & Selection rule and role \\
\midrule
Random sampling & Candidate image IDs and the image-selection seed & Draws uniformly at random from the candidate set. Serves as the reference set before the proposed method replaces images. \\
$k$-center selection on features from a separate model & Image features from a ResNet-50 that is separate from the detector under training & Normalizes the 2,048-dimensional features of a ResNet-50 pretrained on ImageNet-1K and repeatedly selects the image farthest in feature space from the labeled set. Uses neither the outputs nor the internal features of the target detector. \\
Output entropy & Final outputs of the target detector & Aggregates, per image, the binary entropy computed from the top-class confidence of each predicted box. Uses neither internal features nor calibrated error probabilities. \\
Selection by rarity of the internal representation & Internal features of the target detector & Computes the minimum cosine distance between each image's internal representation and those of the images labeled so far, and selects the most distant images. Uses neither output probabilities, error types, nor error probabilities. \\
BADGE adapted to object detection & Internal features and top-class confidence of the target detector & L2-normalizes the mean internal feature of each image, scales it by 1 minus the maximum box confidence, and draws the batch by $k$-means++ seeding on these vectors. Retains the uncertainty and diversity components of BADGE without loss gradients (Methods). Compared as the pre-specified representative of existing methods that use the internal representation. \\
Proposed method & Internal features, outputs, box geometry and neighbor information of the target detector, and a separate 1,500-image labeled set for error estimation & Estimates the error type and probability of each box in the unlabeled candidates and computes an image priority from the effect of each error type, the sufficiency of the evidence and the estimated annotation effort. Starts from the same random draw and replaces, under constraints, only images whose priority difference is confirmed by the stability bounds. \\
\botrule
\end{tabularx}
\end{table}

The external-feature $k$-center selection used ResNet-50 with ImageNet-1K V2 weights and the official image transforms provided by Torchvision \cite{he2016}. This ResNet-50 was separate from the detector under training and was not used to train or evaluate the target detector. Features were generated once per image and stored as 32-bit floating-point values in image-ID order. The distance metric was the squared Euclidean distance after L2 normalization, with ties broken by ascending image ID. The method receives none of the trained weights, internal features, outputs, gradients, predicted boxes, or ground-truth annotations of the target detector.

The output-entropy method computes $-p\log p-(1-p)\log(1-p)$ from the top-class confidence $p$ of each predicted box with confidence of at least 0.05, averages the values within the image, and assigns the maximum value $\log 2$ to an image without such boxes. The same treatment was used at all stages. The BADGE comparator is an adaptation of BADGE \cite{ash2020} to object detection. The original method forms, for each candidate, the gradient of the last-layer loss using the predicted label as a hypothetical label and selects a batch by $k$-means++ seeding on these gradient embeddings. A detector produces a variable number of boxes per image, so we replaced the gradient embedding with an image-level embedding: the mean over the post-NMS boxes of the per-box internal feature vector (the same pooling used for stratification), L2-normalized and multiplied by an image uncertainty of 1 minus the maximum box confidence (1 for an image without detections). The batch is drawn by $k$-means++ seeding. The first image is chosen uniformly at random, and each subsequent image is sampled with probability proportional to the squared distance to the nearest image already selected, with the random state set by the image-selection seed. This adaptation retains the combination of uncertainty and diversity in the target model's representation but does not use loss gradients or hypothetical labels. We therefore do not treat it as a reproduction of the original algorithm.

\subsection*{One cycle of active learning with the proposed method}

The proposed method builds the labeled set in the following stages:

\begin{enumerate}
\item At each stage with 5\%, 10\%, 15\% or 20\% of the candidate images labeled, train the target detector on the current labeled set. Fix the trained detector and run inference on all remaining unlabeled candidate images.
\item For each predicted box, estimate the posterior probabilities of six mutually exclusive outcomes from the target detector's internal features, final outputs, and box geometry: correct, classification error, localization error, combined classification and localization error, duplicate detection, and background false positive.
\item Weight the posterior probability of each error type by that error's estimated effect on the current AP and sum the weighted probabilities within each image. Discount the lower bound of this sum according to the number of correct predictions of the same predicted class obtained from the error-estimation images, and divide by the estimated annotation effort to obtain the image's selection priority.
\item Draw the next 5\% of images uniformly at random as the reference set. Among images with the same predicted classes, predicted object count, predicted object size and similar internal features, match low-priority images inside the reference set with high-priority images outside it. Replace an image only if the priority difference exceeds a fixed threshold after accounting for the stability bounds and if the composition of the selected images does not change excessively. If no image meets the conditions, leave the reference set unchanged.
\item Treat the existing annotations of the selected images as revealed only after the selection is final. Add the selected images to the labeled set and train the detector of the next stage. Repeat until 25\% is reached. An image once selected is not returned to the unlabeled pool.
\end{enumerate}

The proposed method does not simply capture the images with the highest total error probabilities. It considers the effect of each error type, the number of correct predictions of the same predicted class, the expected annotation effort, the composition of the selected images, and the stability of the estimates under box resampling, and reselects only images for which there is evidence to set aside the random draw. Selection uses the predicted boxes retained after nonmaximum suppression (NMS). The confidence threshold for retention was 0.001 for YOLOv8n and 0.05 for Faster R-CNN. The IoU threshold for NMS was 0.7 for YOLOv8n, 0.5 for the ROI head, and 0.7 for the region proposal network of Faster R-CNN. The contributions of at most 100 predicted boxes per image were summed.

\subsection*{Box-level error types}

The proposed method takes as its unit of inference the predicted boxes that the detector outputs after NMS, which suppresses redundant overlapping predictions of the same object and uses at most 100 boxes per image at the time of selection. In calibration images for which labels were available, predicted boxes were matched to ground-truth boxes by TIDE-compatible rules. Each predicted box was assigned to exactly one of six types: correct, classification error, localization error, combined classification and localization error, duplicate detection, or background false positive \cite{bolya2020}. The main analysis used an intersection-over-union (IoU) threshold of 0.5 for matching, where IoU represents the degree of overlap between a predicted box and a ground-truth box. A predicted box whose maximum IoU with any ground-truth box was below 0.1 was classified as a background false positive.

A missed detection, that is, a ground-truth object with no corresponding predicted box, does not have an internal representation associated with a predicted box. Therefore, missed detections were excluded from the calculation of the box-level error probability and evaluated separately based on the number of misses, recall, and TIDE error decomposition. The ``error probability'' in this paper is accordingly defined for individual predicted boxes and does not refer to the probability that an image contains at least one error.

\subsection*{Internal features and error-probability estimation}

For each predicted box $b$, we concatenate $3\times3$ local features obtained by ROI Alignment \cite{he2017mask} from the target detector's multiresolution feature maps, features of an annular region around the box, unnormalized class outputs, top-class confidence, the position and size of the box, and its overlap with and distance from neighboring boxes in the same image. For YOLOv8n, the feature maps are the three inputs of the detection head at strides 8, 16, and 32, ROI-aligned on a $3\times3$ grid at every scale and concatenated. The annular region is obtained by enlarging the box by a factor of 1.5 about its center and is ROI-aligned on a $7\times7$ grid, with the 24 border cells outside the original box retained. The mean and maximum of these cells and the difference between the box-center feature and the ring mean are used. For Faster R-CNN, the internal feature is the 1,024-dimensional output of the box head of the ROI head, and no ring feature is used. All these quantities are associated with the same post-NMS prediction box. Image features extracted by a model other than the detector under training and image metadata were not used for the error probability or the main image selection of the proposed method. The features were standardized and projected onto principal components that explained 95\% of the variance, with a maximum of 64 dimensions fitted to the 750 fit images.

Let $\mathcal{T}$ contain six outcomes: correct, classification error, localization error, combined classification and localization error, duplicate detection, and background false positive. We denote the error type by $t\in\mathcal{T}$, the features by $x_b$ and the predicted class by $c_b$. For each type, we estimated a Gaussian density shared across all classes. Class-specific densities were also estimated when sufficient examples were available. With fewer examples, the estimate decreases continuously toward the shared density:

\begin{equation}
 p(t\mid x_b,c_b)\propto p(\widetilde{x}_b\mid t,c_b)\,\pi_{t,c_b},
 \qquad
 \ell_{t,c}(x)=\lambda_{t,c}\ell^{\mathrm{class}}_{t,c}(x)
 +(1-\lambda_{t,c})\ell^{\mathrm{global}}_{t}(x),
\end{equation}
\begin{equation}
 \lambda_{t,c}=\frac{n_{t,c}}{n_{t,c}+\kappa},
 \qquad \kappa\in\{20,50,100\}.
\end{equation}

To determine $\kappa$, the candidate values 20, 50, and 100 were evaluated using five-fold cross-validation with the fit images split at the image level, and the value that minimized the negative log-likelihood of the held-out images was chosen. Feature standardization and principal component analysis were performed once on all the fit images before choosing $\kappa$. The class priors used Dirichlet smoothing with a strength of 1. Temperature scaling was fitted using the tuning images, which were not included in the evaluation images. For comparison, (i) a model that combines internal features, outputs, and box information and (ii) a model that uses only outputs and box information were fitted using the same splits.

As the additional information carried by the internal features may differ by class and error type, the weight $\rho_{t,c}\in[0,1]$ given to the internal features is obtained by cross-fitting on the fit images. With five image-grouped folds, both models are refit on four folds and scored on the held-out fold, so that every fit box is scored once out of fold. The weight is $\rho_{t,c}=\mathrm{clip}[P_{t,c}\,(1-\mathrm{ECE}_{t,c})\,n_{t,c}/(n_{t,c}+30),\,0,\,1]$, where $P_{t,c}$ is the fraction of 2,000 image-level bootstrap resamples of the out-of-fold scores in which the normalized AUPRC of the model with internal features exceeds that of the model with outputs and box information for type $t$ and predicted class $c$, $\mathrm{ECE}_{t,c}$ is the adaptive expected calibration error of the model with internal features with 10 equal-count bins, and $n_{t,c}$ is the number of fit boxes of type $t$ and class $c$. Cells with fewer than 20 boxes use the value computed with all classes of type $t$ pooled. No parameter used to calculate $\rho$ was tuned. After temperature scaling, probabilities below $10^{-6}$ were set to $10^{-6}$, and each model's posterior was renormalized to sum to 1 across the six outcomes. The posteriors of the model with internal features and those of the model with outputs and box information were mixed as

\begin{equation}
 \widetilde{p}_{t,b}=\rho_{t,c}p^{\mathrm{fused}}_{t,b}
 +(1-\rho_{t,c})p^{\mathrm{output}}_{t,b},
 \qquad
 p^{*}_{t,b}=\frac{\widetilde{p}_{t,b}}{\sum_{t'\in\mathcal{T}}\widetilde{p}_{t',b}}.
\end{equation}

As $\rho_{t,c}$ differs between error types, the mixture $\widetilde{p}_{t,b}$ need not sum to 1 over the types. Therefore, we normalize the mixture for each box by the sum over types and take $p^{*}_{t,b}$ as the final posterior. The error probability defined below and the priority described later use this normalized value. For error types and predicted classes for which the evidence for improvement from internal features is weak, $\rho_{t,c}$ is small, and the estimate approaches that obtained using outputs and box information only. An image is replaced only when the priority difference and compositional constraints described later are satisfied. Otherwise, a random draw is retained. The error probability of box $b$ is defined as

\begin{equation}
 p^{\mathrm{error}}_b
 =1-p^{*}_{\mathrm{correct},b}
 =\sum_{t\ne\mathrm{correct}}p^{*}_{t,b}.
\end{equation}

For the 375 evaluation images, we reported four metrics: the prevalence-corrected AUPRC, Brier score, negative log-likelihood, and adaptive expected calibration error, each computed per error type and averaged across the five error types. The prevalence-corrected AUPRC is $(\mathrm{AUPRC}-\pi_t)/(1-\pi_t)$, where $\pi_t$ is the fraction of evaluation boxes of type $t$. An error type with fewer than five positive boxes among the evaluation boxes was excluded from the average. This evaluation assessed the posterior of the model with internal features, $p^{\mathrm{fused}}_{t,b}$, and the posterior of the model with outputs and box information only, $p^{\mathrm{output}}_{t,b}$. The mixed posterior $p^{*}_{t,b}$ was not evaluated (Fig.~\ref{fig:error-ranking}). AUPRC measures how well the predictions that are likely to be errors are ranked at the top. The other three metrics assess the accuracy of the predicted probabilities. Image-level resampling (200 resamples per error type, with all boxes of a resampled image taken together) provided a one-sided 90\% lower bound for the AUPRC difference between a score with internal features and a score with outputs and box information only. The reported lower bound for the average over types is the mean of the per-type lower bounds. Because the same evaluation images were reused over four cycles, the cycles were not treated as independent replicates. These evaluation metrics were distinct from the primary measure of performance improvement obtained from the added data.

$p^{\mathrm{error}}_b$ describes the total error probability of a box and is used to examine its relationship with output uncertainty. Image selection does not use this sum directly, because the weights $w_t$ introduced below differ between error types, and the summed $p^{\mathrm{error}}_b$ cannot recover the expected effect of an image's errors. Selection priority uses each incorrect $p^{*}_{t,b}$.

\subsection*{Image priority reflecting the effect of errors on detection performance}

The number of errors did not match their effect on mAP. Therefore, using the 750 fit images, we defined the weight for each error type as the change in average precision, $\Delta AP_t$, which would follow if only the errors of type $t$ were corrected, divided by the number of such errors $N_t$. This AP was the binary AP at an IoU threshold of 0.5, calculated over the pooled predictions of the fit images and expressed on a scale from 0 to 1. Classification, localization, and combined errors are corrected by treating them as true positives, and duplicate detections and background false positives are corrected by removing them. The replacement margin $\delta$ introduced below is defined on this scale. This weight is the estimated effect of that error on the current evaluation metric. It does not directly express the learnability of an image or the performance gain from training. Specifically,

\begin{equation}
 w_t=\frac{\max(0,\Delta AP_t)}{N_t+\varepsilon}
\end{equation}

with $\varepsilon=1$. The weight was averaged across the folds of a five-fold split performed at the image level. For error types with fewer than 20 instances across all folds, the weight was replaced with the mean weight of the types with sufficient examples. The expected effect of errors in image $i$ is

\begin{equation}
 B_i=\sum_{b\in D_i}\sum_{t\ne\mathrm{correct}}w_t p^{*}_{t,b},
\end{equation}

where $D_i$ is the set of post-NMS predicted boxes that pass the confidence threshold described in the previous section. $B_i$ is the weighted sum of the expected error counts, and not the probability that an image contains at least one error. The evidence sufficiency $q_i$ of an image is calculated using the number of correct examples $n_{\mathrm{correct},c_b}$ for the predicted class $c_b$ of each box $b$:
\begin{equation}
 q_i=\frac{1}{|D_i|}\sum_{b\in D_i}
 \frac{n_{\mathrm{correct},c_b}}{n_{\mathrm{correct},c_b}+\kappa},
\end{equation}
where $q_i=0$ when $D_i$ is empty. As $q_i$ is small when few correct predictions of the same predicted class are available, images with less supporting evidence receive a lower priority.

The estimated annotation effort uses the number of predicted boxes $|D_i|$ and the crowding of boxes of the same predicted class:
\begin{equation}
 \widehat C_i=\max\!\left(10^{-6},\,1+|D_i|+\sum_{b\in D_i}n_b\right),
 \qquad
 n_b=\sum_{\substack{b'\in D_i,\ b'\ne b\\c_{b'}=c_b}}
 \frac{1}{1+d(b,b')},
\end{equation}
where $d(b,b')$ is the distance between the box centers, normalized by the image diagonal. The coefficients of the constant, box count, and crowding terms were fixed at 1.

The predicted boxes of an image were resampled 1,000 times with replacement, and the 2.5th and 97.5th percentiles of $B_i$ were used as the lower bound $B_i^{\mathrm{LCB}}$ and the upper bound $B_i^{\mathrm{UCB}}$, respectively, with the random seed fixed at 0. Using $q_i$, which is based on the number of correct predictions of the same predicted class, and the estimated effort $\widehat C_i$, we define

\begin{equation}
 A_i=A_i^{\mathrm{LCB}}
 =\frac{q_i B_i^{\mathrm{LCB}}}{\widehat C_i},
 \qquad
 A_i^{\mathrm{UCB}}
 =\frac{q_i B_i^{\mathrm{UCB}}}{\widehat C_i}
\end{equation}

where $A_i$ is the selection priority and $A_i^{\mathrm{UCB}}$ is its upper bound. These bounds quantify the sensitivity of the score to resampling the predicted boxes within an image, conditional on the fitted error estimators and error-type weights. They are heuristic stability bounds rather than calibrated confidence intervals for the true error burden or for the gain from retraining. For example, an image with a single predicted box has identical bounds even though the error probability of that box remains uncertain. $A_i$ uses the lower bound so that images whose scores are unstable under box resampling do not receive an increased priority. $\widehat C_i$ is estimated from a fixed per-image effort, number of predicted boxes, and their crowding. The true number of ground-truth boxes was not used for selection. Images without detected boxes have $B_i=0$ but remain eligible through random sampling. $A_i$ is an index for prioritizing images that are predicted to contain errors, adjusted by a relative value computed from the number and crowding of predicted boxes. It does not represent the human labeling time.

\subsection*{Replacement between images with matching characteristics}

For each image-selection seed, the random selection and the proposed method were initialized from the same uniformly random set of images $R$. Candidate images were stratified by the set of predicted classes, the number of predicted boxes in bins of 0, 1 to 2, 3 to 5, 6 to 10, and 11 or more, the dominant predicted object size in the COCO small, medium, and large bins (areas below $32^2$, from $32^2$ to below $96^2$, and $96^2$ or above, respectively), and one of eight $k$-means clusters formed from the standardized mean internal feature of each image (random state 0, 10 initializations). A low-priority image $i$ inside $R$ is matched with a high-priority image $j$ outside $R$ in the same stratum. Image $i$ is replaced by $j$ only if $A_j^{\mathrm{LCB}}>A_i^{\mathrm{UCB}}$ and the evidence margin $A_j^{\mathrm{LCB}}-A_i^{\mathrm{UCB}}$ exceeds a prespecified $\delta$. None of the images were used more than once. Among the candidates with the same evidence margin, the image farthest from the already selected images in the internal-feature space was selected first.

At most 50\% of the random draw was replaced. Replacements that would change the marginal distributions of the predicted class, predicted object size, or number of predicted boxes beyond tolerance were not accepted. The change was measured by the Jensen--Shannon divergence. Images without predicted boxes were used neither as replacements nor as images to be replaced and remained in the selection only when they were included in the random draw. The share of images with zero or one predicted box at a confidence of 0.25 or above was kept from falling below their share in the random draw. In addition, the proportions of classes of the dominant-size and object-count bins, computed from all predicted boxes retained for selection, had to remain within the central 95\% range, that is, between the 2.5th and 97.5th percentiles, of 1,000 resamples of the random draw. When the evidence for replacement was insufficient, a random draw was retained. These constraints prevented the selection from concentrating on particular groups of images and allowed the contribution of internal error information to be compared under matched conditions.

In the experiments, the evidence margin required for replacement was fixed at $\delta=1.35\times10^{-7}$, and the upper limit of the Jensen--Shannon divergence was set at 0.00561. The Jensen--Shannon divergence and shares of images with zero or one box were checked on boxes with a confidence of 0.25 or above, whereas the resampled composition range was calculated using all boxes retained for selection.

\subsection*{Additional experiment that changed only the selection criterion under matched conditions}

To examine the difference between linking the internal representation to errors and using representation rarity alone, we ran a single comparison from the common model trained on 10\% of the labeled data to the 15\% labeled-data level, with image-selection seed 11 and model-training seed 1. The starting model, candidate set, maximum number of replaced images, stratification, constraints on the selection distribution, model-training seed, optimizer, data augmentation, number of training epochs, and evaluation method were identical. Only the selection priority was changed. Four criteria were compared, including that of the proposed method. The first is the proposed method, which uses the error probability $p^{*}_{t,b}$. The method combines internal features, outputs, and box information. The second, ``outputs and box information only,'' does not use internal features for error-probability estimation. It uses the posterior $p^{\mathrm{output}}_{t,b}$ of the outputs-and-box model described above in place of $p^{*}_{t,b}$ (implementation name \texttt{output\_matched\_replace}). The per-type weights $w_t$, evidence sufficiency $q_i$, estimated effort $\widehat C_i$, and stability bounds of the priority, stratification, and replacement rules are identical to those of the proposed method. The output-entropy selection of the main analysis differs from this criterion in that it aggregates the binary entropy of the top-class confidence within an image and selects directly from the top without error types, weights, or replacements. Third, under the ``internal features shuffled across images'' criterion, the internal features of each predicted box are replaced at random with those of another box of the same predicted class and the same predicted size bin. The shuffle is applied independently to the fit, tuning, evaluation, and candidate splits using random numbers derived deterministically from the dataset, seeds, split, and stage. The outputs, box information, neighbor information, and assignment of boxes to images are held fixed. The error-probability model, temperature-scaling parameter, and $\rho_{t,c}$ are then re-estimated on the shuffled features, and images are selected with the same weighting and replacement rules as the proposed method (implementation name \texttt{shuffled\_internal\_matched\_replace}). This criterion controls for the contribution of image-specific information carried by the internal features. The strata used for matching were defined from the unshuffled internal features and shared by all four criteria. Fourth, selection by the rarity of the internal representation alone uses no error labels, per-type posteriors, or error probabilities. It only checks how far a candidate image's internal features lie from the labeled set (implementation name \texttt{generic\_internal\_novelty\_matched\_replace}).

The primary measure for comparing the selection rules was the difference in mAP@[0.50:0.95] on the official validation set at the matched 15\% point. The rate at which the selected images actually contained errors was a secondary measure computed after selection using the ground-truth annotations. It was not used to tune priorities or thresholds, or to reselect images for any comparator method.

\subsection*{Verification of selection behavior from sealed records}

For the proposed method and BADGE, we retrieved the selected image IDs and selection summaries saved at every cycle for two conditions: image-selection seed 11 with model-training seed 1 and image-selection seed 18 with model-training seed 0. There were eight experiments (two datasets, two seed conditions, and two methods), and for each experiment, the selected image IDs and summaries of cycles one to four amounted to 64 artifacts in total. The SHA-256 hash and size of each file were checked against the artifact manifest sealed when the experiment concluded and against the published records.

For the proposed method, we tallied the number of images actually replaced from the random draw at each stage, the number of candidates rejected because their margins were below the threshold, and the replacement rate. The overlap with the selections of BADGE was reported as the number of shared images and the Jaccard index. Because the two methods have different parent models after retraining, this comparison describes their selection behavior and does not isolate the effect of error information alone.

\subsection*{Experiment matrix and observation points}

The main analysis comprised 24 learning curves: YOLOv8n, two datasets, two seed conditions, and six methods. The supplementary Faster R-CNN comparison comprised six learning curves: two datasets, image-selection seed 11 with model-training seed 1, and three methods. The 40-epoch supplementary experiment comprised five prespecified learning curves on BDD100K and one additional learning curve for rarity-only selection. Each learning curve had five observation points at 5, 10, 15, 20, and 25\% labeled data, and all methods within a dataset, detector, and seed condition shared the same model trained on 5\% of the labeled data. The matched additional experiment, which changed only the selection criterion, started from the common model trained on 10\% of the labeled data and extended to the 15\% labeled-data level, with image-selection seed 11 and model-training seed 1. The configuration, selected images, trained models, and run history of each curve were fixed, and the official validation set was evaluated using a single evaluation implementation.

\subsection*{Evaluation metrics}

The detection performance was measured mainly using mAP@[0.50:0.95] across all official evaluation images of each dataset. AP50, AP75, per-class AP, AP by object size, precision, recall, and TIDE error breakdown were also examined. The primary effect size for comparison between methods was the normalized area under the learning curve: mAP@[0.50:0.95] as a function of the labeled fraction $x\in\{0.05,0.10,0.15,0.20,0.25\}$, integrated using the trapezoidal rule and divided by the 0.20 width of the range. This value corresponds to the mean mAP from 5\% to 25\% labeled data and is not calculated per selected image.

\begin{equation}
 \mathrm{AULC}_{\mathrm{norm}}
 =\frac{1}{0.20}\sum_{j=1}^{4}
 \frac{m(x_j)+m(x_{j+1})}{2}(x_{j+1}-x_j).
\end{equation}

The selection behavior was described by the number of replacements from the sealed artifacts, the replacement rate per stage, and the Jaccard index between the selected sets of the proposed method and BADGE. The number of true errors per 100 images was also reported in the matched comparison at the 15\% labeled-data level with image-selection seed 11 and model-training seed 1, which changed only the selection criterion.

\subsection*{Statistics and reproducibility}

\subsubsection*{Statistical analyses}

For each detector, dataset, and pair of image-selection and model-training seeds, the mAP at 5, 10, 15, 20, and 25\% labeled data was integrated using the trapezoidal rule and divided by the 0.20 width of the range to obtain the normalized area under the learning curve. In the YOLOv8n main analysis, the differences between the proposed method and rarity-only selection, BADGE, output-entropy selection, external-feature $k$-center selection, and random sampling were computed using the matched dataset, image-selection seed, model-training seed, and evaluation images. For the five comparisons under each of the four conditions, 95\% intervals were obtained by resampling the evaluation images. In the supplementary Faster R-CNN comparison, the differences of the proposed method from BADGE and rarity-only selection were similarly computed for the two datasets. In a sensitivity analysis that excluded the shared starting point, the mean of the learning curve and the differences between the methods were computed over the four points at 10\%, 15\%, 20\%, and 25\%, excluding the 5\% point.

To express the variation from changing the evaluation images with the trained models fixed, the same images were matched across methods and resampled 10,000 times with resampling seed 20260728, and the 2.5th and 97.5th percentiles of the difference formed the 95\% interval. Each resampled image set was shared by all methods and all labeled fractions, and the AP of each resample was recomputed from the per-image matching records of the official evaluation rather than from per-image AP values. These intervals do not include variation arising from repeated training or image selection with different image-selection or model-training seeds. The two seed conditions were reported separately, and we did not use them to estimate the confidence intervals or rank probabilities over other seeds.

For the additional experiment at the 15\% labeled-data level, which changed only the selection criterion, differences were computed using the same official evaluation images with 95\% intervals from 10,000 resamples in which the evaluation images were matched across criteria. Analyses by class and error type were used to describe the breakdown of the overall performance difference and are not treated as standalone tests of superiority. Each combination of the detector, dataset, seed pair, and selection method was trained once. The only replication over random numbers was in the two conditions with different image-selection and model-training seeds. No further replicates were run. The official evaluation set contains 5,000 images from COCO 2017 and 10,000 images from BDD100K. No null hypothesis tests were performed, and the intervals were used descriptively. Therefore, no correction for multiple comparisons was applied.

\subsection*{Label efficiency and computational efficiency}

The methods were compared for the same labeled fractions. Because comparable human labeling and GPU times were not obtained across methods, reductions in labeling or computing time were not evaluated as primary outcomes. The 1,500 images that the proposed method uses for error probability estimation were treated as development resources, separate from the labeled set used for selection.

\subsection*{Missing data and run history}

The six-method comparison was not performed for a YOLOv8n dataset and seed condition if either the learning curve or the official evaluation for any of the six methods were missing. The supplementary Faster R-CNN comparison required all three methods for a dataset. Missing values were neither interpolated nor extrapolated. Runs resumed from a saved training stage were treated as continuations of the original experiment, after the identity of the saved stage had been verified using a hash. The final comparisons used only learning curves in which the SHA-256 hashes of the model weights and labeled set for the model trained on 5\% of the labeled data matched across the compared methods within each dataset, detector, and seed condition.

The SHA-256 hashes of the configuration and inputs, initialization record, selected image IDs, random reference draw, replaced image pairs, candidate set at each stage, trained models, epoch history, evaluation results, GPU time, and execution environment were recorded for each experimental unit.

\subsection*{Use of generative AI}

Generative AI tools, specifically OpenAI Codex and Anthropic Claude, assisted with translation, language editing, software implementation, and the scripts used to prepare the figures. The authors reviewed and verified all outputs and remain responsible for the research question, experimental design, implementation, data analysis, citations, and manuscript. No generative AI tool has been listed as an author.

\backmatter

\bmhead{Data availability}

COCO 2017 was obtained from the official download page (\url{https://cocodataset.org/#download}): 118,287 train2017 images, 5,000 val2017 images, and 2017 Train/Val annotations. BDD100K was obtained from the official Berkeley DeepDrive data portal (\url{https://bdd-data.berkeley.edu/download.html}): 70,000 training images and 10,000 validation images from 100K Images, together with Detection 2020 Labels. Both datasets were used under the terms set by the providers.

The following annotation files that were fixed for the experiments and their SHA-256 values are:
\begin{itemize}
  \item COCO 2017, \path{instances_train2017.json}\\ {\footnotesize\ttfamily 610fce4944abdeb15354cc765333805529359d12d88f2f711393ca586901d01d}
  \item COCO 2017, \path{instances_val2017.json}\\ {\footnotesize\ttfamily e8c7f7908f1d7278341fae127d0da654f102f11bd7b21d8aeefa635b8c810b6f}
  \item BDD100K, \path{det_train.json}\\ {\footnotesize\ttfamily 2bd0068966b516d6d1f2fef7c8d6d90f07f76fe346ba170de52faac03239a46f}
  \item BDD100K, \path{det_val.json}\\ {\footnotesize\ttfamily 1a1b42ff519fce67b03daf41c0122c81bc31452e333b7ccb956a0a5fd72ac04c}
\end{itemize}

For each of the image-selection seeds 11 and 18 used in the reported comparisons, an \texttt{ID\_LISTS.json} file that recorded the image IDs for COCO 2017 and the image file names for BDD100K was fixed. Each file records the role of each image: the labeling candidate, one of the 1,500 calibration images, or one of the 1,000 internal-validation images. The labeled candidates numbered 115,787 for COCO 2017 and 67,363 for BDD100K. The image ID lists, their SHA-256 values, the Source Data corresponding to the analyses, and the minimal data needed to verify the results are included in Supplementary Software 1, which was provided to the journal for peer review and will be deposited in a permanent public repository upon acceptance. The images from the third-party datasets were not redistributed. The values shown in Figs.~2--5 and Supplementary Fig.~1 are provided as Source Data. The Supplementary Information accompanies this preprint as an ancillary file.

\bmhead{Code availability}

The training and analysis code written for this study (ELIS, Error-Linked Image Selection), the fixed study plan, the analysis plan, the method registry, the experimental matrix, the environment setup instructions, and the figure-regeneration scripts are included in Supplementary Software 1, which was provided to the journal for peer review and will be deposited in a permanent public repository with a versioned release upon acceptance. Supplementary Software 1 also includes an implementation that reruns the proposed method and comparators, a minimal dataset comprising selected image IDs, per-run evaluation values, and Source Data for the figures. The code will be released under the PolyForm Noncommercial License 1.0.0, which permits use, modification, and redistribution for noncommercial purposes, including reproduction and extension of the reported results. Commercial use requires a separate license from Adansons Corp. Third-party software, including Ultralytics 8.4.101, PyTorch 2.13.0, and torchvision 0.28.0, remains under its own license, and the versions of all packages used are listed in Supplementary Software 1.

\bmhead{Funding}

This study was supported by JSPS KAKENHI Grant Number JP26K05106.

\bmhead{Author contributions}

Y.N. and S.I. conceived the research questions and designed the study. Y.N. developed the method, designed and conducted the experiments, and analyzed and interpreted the results. Y.N. and K.H. developed the core algorithm, and K.H. implemented the proposed method and the experimental software. Y.N. wrote the manuscript and S.I. critically reviewed and revised it. All authors discussed the results and approved the final manuscript.

\bmhead{Competing interests}

Y.N. and K.H. are affiliated with Adansons Corp. The findings may be integrated into future Adansons products. Adansons Corp. holds the copyright in the code written for this study and offers commercial licenses for it. The authors declare no other competing interests.

\clearpage


\begin{thebibliography}{99}

\bibitem{sambasivan2021}
Sambasivan, N. \textit{et al.} ``Everyone wants to do the model work, not the data work'': data cascades in high-stakes AI. In \textit{Proc. 2021 CHI Conf. Human Factors Comput. Syst.} Article 39, 1--15 (ACM, 2021). \href{https://doi.org/10.1145/3411764.3445518}{doi:10.1145/3411764.3445518}.

\bibitem{bernhardt2022}
Bernhardt, M. \textit{et al.} Active label cleaning for improved dataset quality under resource constraints. \textit{Nat. Commun.} \textbf{13}, 1161 (2022). \href{https://doi.org/10.1038/s41467-022-28818-3}{doi:10.1038/s41467-022-28818-3}.

\bibitem{geuenich2024}
Geuenich, M. J., Gong, D.-W. \& Campbell, K. R. The impacts of active and self-supervised learning on efficient annotation of single-cell expression data. \textit{Nat. Commun.} \textbf{15}, 1014 (2024). \href{https://doi.org/10.1038/s41467-024-45198-y}{doi:10.1038/s41467-024-45198-y}.

\bibitem{whang2023}
Whang, S. E., Roh, Y., Song, H. \& Lee, J.-G. Data collection and quality challenges in deep learning: a data-centric AI perspective. \textit{VLDB J.} \textbf{32}, 791--813 (2023). \href{https://doi.org/10.1007/s00778-022-00775-9}{doi:10.1007/s00778-022-00775-9}.

\bibitem{mazumder2023}
Mazumder, M. \textit{et al.} DataPerf: benchmarks for data-centric AI development. \textit{Adv. Neural Inf. Process. Syst.} \textbf{36}, 5320--5347 (2023). \href{https://doi.org/10.52202/075280-0235}{doi:10.52202/075280-0235}.

\bibitem{mindermann2022}
Mindermann, S. \textit{et al.} Prioritized training on points that are learnable, worth learning, and not yet learnt. \textit{Proc. Mach. Learn. Res.} \textbf{162}, 15630--15649 (2022). \href{https://proceedings.mlr.press/v162/mindermann22a.html}{Proceedings of Machine Learning Research}.

\bibitem{hacohen2022}
Hacohen, G., Dekel, A. \& Weinshall, D. Active learning on a budget: opposite strategies suit high and low budgets. \textit{Proc. Mach. Learn. Res.} \textbf{162}, 8175--8195 (2022). \href{https://proceedings.mlr.press/v162/hacohen22a.html}{Proceedings of Machine Learning Research}.

\bibitem{settles2009}
Settles, B. \textit{Active Learning Literature Survey}. Computer Sciences Technical Report 1648 (University of Wisconsin--Madison, 2009). \href{https://research.cs.wisc.edu/techreports/2009/TR1648.pdf}{University of Wisconsin--Madison}.

\bibitem{ren2021survey}
Ren, P. \textit{et al.} A survey of deep active learning. \textit{ACM Comput. Surv.} \textbf{54}, Article 180, 1--40 (2022). \href{https://doi.org/10.1145/3472291}{doi:10.1145/3472291}.

\bibitem{gal2017}
Gal, Y., Islam, R. \& Ghahramani, Z. Deep Bayesian active learning with image data. \textit{Proc. Mach. Learn. Res.} \textbf{70}, 1183--1192 (2017). \href{https://proceedings.mlr.press/v70/gal17a.html}{Proceedings of Machine Learning Research}.

\bibitem{sener2018}
Sener, O. \& Savarese, S. Active learning for convolutional neural networks: a core-set approach. In \textit{International Conference on Learning Representations} (2018). \href{https://openreview.net/forum?id=H1aIuk-RW}{OpenReview}.

\bibitem{ash2020}
Ash, J. T., Zhang, C., Krishnamurthy, A., Langford, J. \& Agarwal, A. Deep batch active learning by diverse, uncertain gradient lower bounds. In \textit{International Conference on Learning Representations} (2020). \href{https://openreview.net/forum?id=ryghZJBKPS}{OpenReview}.

\bibitem{choi2021}
Choi, J., Elezi, I., Lee, H.-J., Farabet, C. \& Alvarez, J. M. Active learning for deep object detection via probabilistic modeling. In \textit{Proc. IEEE/CVF Int. Conf. Comput. Vis.} 10244--10253 (2021). \href{https://doi.org/10.1109/ICCV48922.2021.01010}{doi:10.1109/ICCV48922.2021.01010}.

\bibitem{wu2022}
Wu, J., Chen, J. \& Huang, D. Entropy-based active learning for object detection with progressive diversity constraint. In \textit{Proc. IEEE/CVF Conf. Comput. Vis. Pattern Recognit.} 9387--9396 (2022). \href{https://doi.org/10.1109/CVPR52688.2022.00918}{doi:10.1109/CVPR52688.2022.00918}.

\bibitem{yu2022cald}
Yu, W., Zhu, S., Yang, T. \& Chen, C. Consistency-based active learning for object detection. In \textit{Proc. IEEE/CVF Conf. Comput. Vis. Pattern Recognit. Workshops} 3950--3959 (2022). \href{https://doi.org/10.1109/CVPRW56347.2022.00440}{doi:10.1109/CVPRW56347.2022.00440}.

\bibitem{yang2024}
Yang, C., Huang, L. \& Crowley, E. J. Plug and play active learning for object detection. In \textit{Proc. IEEE/CVF Conf. Comput. Vis. Pattern Recognit.} 17784--17793 (2024). \href{https://doi.org/10.1109/CVPR52733.2024.01684}{doi:10.1109/CVPR52733.2024.01684}.

\bibitem{munjal2022}
Munjal, P., Hayat, N., Hayat, M., Sourati, J. \& Khan, S. Towards robust and reproducible active learning using neural networks. In \textit{Proc. IEEE/CVF Conf. Comput. Vis. Pattern Recognit.} 223--232 (2022). \href{https://doi.org/10.1109/CVPR52688.2022.00032}{doi:10.1109/CVPR52688.2022.00032}.

\bibitem{yehuda2022}
Yehuda, O., Dekel, A., Hacohen, G. \& Weinshall, D. Active learning through a covering lens. \textit{Adv. Neural Inf. Process. Syst.} \textbf{35}, 22354--22367 (2022). \href{https://doi.org/10.52202/068431-1624}{doi:10.52202/068431-1624}.

\bibitem{liang2024aide}
Liang, M. \textit{et al.} AIDE: an automatic data engine for object detection in autonomous driving. In \textit{Proc. IEEE/CVF Conf. Comput. Vis. Pattern Recognit.} 14695--14706 (2024). \href{https://doi.org/10.1109/CVPR52733.2024.01392}{doi:10.1109/CVPR52733.2024.01392}.

\bibitem{bolya2020}
Bolya, D., Foley, S., Hays, J. \& Hoffman, J. TIDE: a general toolbox for identifying object detection errors. In \textit{Computer Vision--ECCV 2020} 558--573 (Springer, 2020). \href{https://doi.org/10.1007/978-3-030-58580-8_33}{doi:10.1007/978-3-030-58580-8\_33}.

\bibitem{guo2017}
Guo, C., Pleiss, G., Sun, Y. \& Weinberger, K. Q. On calibration of modern neural networks. \textit{Proc. Mach. Learn. Res.} \textbf{70}, 1321--1330 (2017). \href{https://proceedings.mlr.press/v70/guo17a.html}{Proceedings of Machine Learning Research}.

\bibitem{kuppers2020}
K\"uppers, F., Kronenberger, J., Shantia, A. \& Haselhoff, A. Multivariate confidence calibration for object detection. In \textit{Proc. IEEE/CVF Conf. Comput. Vis. Pattern Recognit. Workshops} 326--327 (2020). \href{https://openaccess.thecvf.com/content_CVPRW_2020/html/w20/Kuppers_Multivariate_Confidence_Calibration_for_Object_Detection_CVPRW_2020_paper.html}{Computer Vision Foundation}.

\bibitem{schubert2021}
Schubert, M., Kahl, K. \& Rottmann, M. MetaDetect: uncertainty quantification and prediction quality estimates for object detection. In \textit{2021 Int. Joint Conf. Neural Netw.} 1--10 (IEEE, 2021). \href{https://doi.org/10.1109/IJCNN52387.2021.9534289}{doi:10.1109/IJCNN52387.2021.9534289}.

\bibitem{jocher2023}
Jocher, G., Chaurasia, A. \& Qiu, J. \textit{Ultralytics YOLOv8} (version 8.0.0) (2023). \href{https://github.com/ultralytics/ultralytics}{GitHub repository}.

\bibitem{lin2014}
Lin, T.-Y. \textit{et al.} Microsoft COCO: common objects in context. In \textit{Computer Vision--ECCV 2014} 740--755 (Springer, 2014). \href{https://doi.org/10.1007/978-3-319-10602-1_48}{doi:10.1007/978-3-319-10602-1\_48}.

\bibitem{yu2020bdd}
Yu, F. \textit{et al.} BDD100K: a diverse driving dataset for heterogeneous multitask learning. In \textit{Proc. IEEE/CVF Conf. Comput. Vis. Pattern Recognit.} 2633--2642 (2020). \href{https://doi.org/10.1109/CVPR42600.2020.00271}{doi:10.1109/CVPR42600.2020.00271}.

\bibitem{he2016}
He, K., Zhang, X., Ren, S. \& Sun, J. Deep residual learning for image recognition. In \textit{Proc. IEEE Conf. Comput. Vis. Pattern Recognit.} 770--778 (2016). \href{https://doi.org/10.1109/CVPR.2016.90}{doi:10.1109/CVPR.2016.90}.

\bibitem{he2017mask}
He, K., Gkioxari, G., Doll\'ar, P. \& Girshick, R. Mask R-CNN. In \textit{Proc. IEEE Int. Conf. Comput. Vis.} 2961--2969 (2017). \href{https://openaccess.thecvf.com/content_iccv_2017/html/He_Mask_R-CNN_ICCV_2017_paper.html}{Computer Vision Foundation}.

\end{thebibliography}
\end{document}